\documentclass[letterpaper]{article}
\usepackage{aaai24} 
\usepackage{times} 
\usepackage{helvet} 
\usepackage{courier} 
\usepackage[hyphens]{url} 
\usepackage{graphicx} 
\usepackage{graphicx} 
\usepackage{natbib} 
\usepackage{caption} 
\usepackage{algorithm}
\usepackage{algorithmic}

\usepackage{pdfpages}

\title{Mitigating Label Noise through Data Ambiguation} 

\author{
Julian~Lienen\textsuperscript{\rm 1},
Eyke~H\"ullermeier\textsuperscript{\rm 2,\rm 3}
}
\affiliations {
\textsuperscript{\rm 1}Department of Computer Science, Paderborn University\\
\textsuperscript{\rm 2}Institute of Informatics, LMU Munich \\
\textsuperscript{\rm 3}Munich Center for Machine Learning \\
julian.lienen@upb.de,
eyke@lmu.de
}

\usepackage{bibentry}
\usepackage{booktabs}       

\usepackage{amsmath}
\usepackage{amsfonts}
\usepackage{multirow,makecell}
\usepackage{amssymb}
\usepackage{pifont}

\usepackage{subcaption}

\newcommand{\pname}{Robust Data Ambiguation}
\newcommand{\pnames}{RDA}

\newcommand{\R}{{\mathbb{R}}}

\renewcommand{\P}{{\mathbb{P}}}

\newcommand{\cX}{\mathcal{X}}
\newcommand{\cY}{\mathcal{Y}}

\newcommand{\argmax}{\operatorname*{argmax}}

\newcommand{\fromto}{\longrightarrow}

\newcommand*{\defeq}{\mathrel{\vcenter{\baselineskip0.5ex \lineskiplimit0pt
			\hbox{\footnotesize.}\hbox{\footnotesize.}}}%
	=}

\newcommand{\hatp}{\widehat{p}}

\newcommand{\given}{\, | \,}

\renewcommand{\vec}[1]{\boldsymbol{#1}}

\newcommand{\cmark}{\ding{51}}%
\newcommand{\xmark}{\ding{55}}%

\begin{document}
\maketitle

\begin{abstract}
    Label noise poses an important challenge in machine learning, especially in deep learning, in which large models with high expressive power dominate the field. Models of that kind are prone to memorizing incorrect labels, thereby harming generalization performance. Many methods have been proposed to address this problem, including robust loss functions and more complex label correction approaches. Robust loss functions are appealing due to their simplicity, but typically lack flexibility, while label correction usually adds substantial complexity to the training setup.
    In this paper, we suggest to address the shortcomings of both methodologies by ``ambiguating'' the target information, adding additional, complementary candidate labels in case the learner is not sufficiently convinced of the observed training label. More precisely, we leverage the framework of so-called superset learning to construct set-valued targets based on a confidence threshold, which deliver imprecise yet more reliable beliefs about the ground-truth, effectively helping the learner to suppress the memorization effect. In an extensive empirical evaluation, our method demonstrates favorable learning behavior on synthetic and real-world noise, confirming the effectiveness in detecting and correcting erroneous training labels.
\end{abstract}

\section{Introduction}\label{sec:intro}

Label noise refers to the presence of incorrect or unreliable annotations in the training data, which can negatively impact the generalization performance of learning methods. Dealing with label noise constitutes a major challenge for the application of machine learning methods to real-world applications, which often exhibit noisy annotations in the form of erroneous labels or distorted sensor values. This is no less a concern for large models in the regime of deep learning, which have become increasingly popular due to their high expressive power, and which are not immune to the harming nature of label noise either. Existing methods addressing this issue include off-the-shelf robust loss functions, e.g., as proposed by \citet{Wang2019SymmetricCE}, and label correction, for example by means of replacing labels assumed to be corrupted \citep{DBLP:conf/aaai/WuSX0M21}. While the former appeals with its effortless integration into classical supervised learning setups, methods of the latter kind allow for a more effective suppression of noise by remodeling the labels. However, this comes at the cost of an increased model complexity, typically reducing the efficiency of the training \citep{Liu2020EarlyLearningRP}.

\begin{figure}[t]
    \centering
    \begin{subfigure}[b]{0.49\linewidth}
        \centering
        \includegraphics[width=\textwidth]{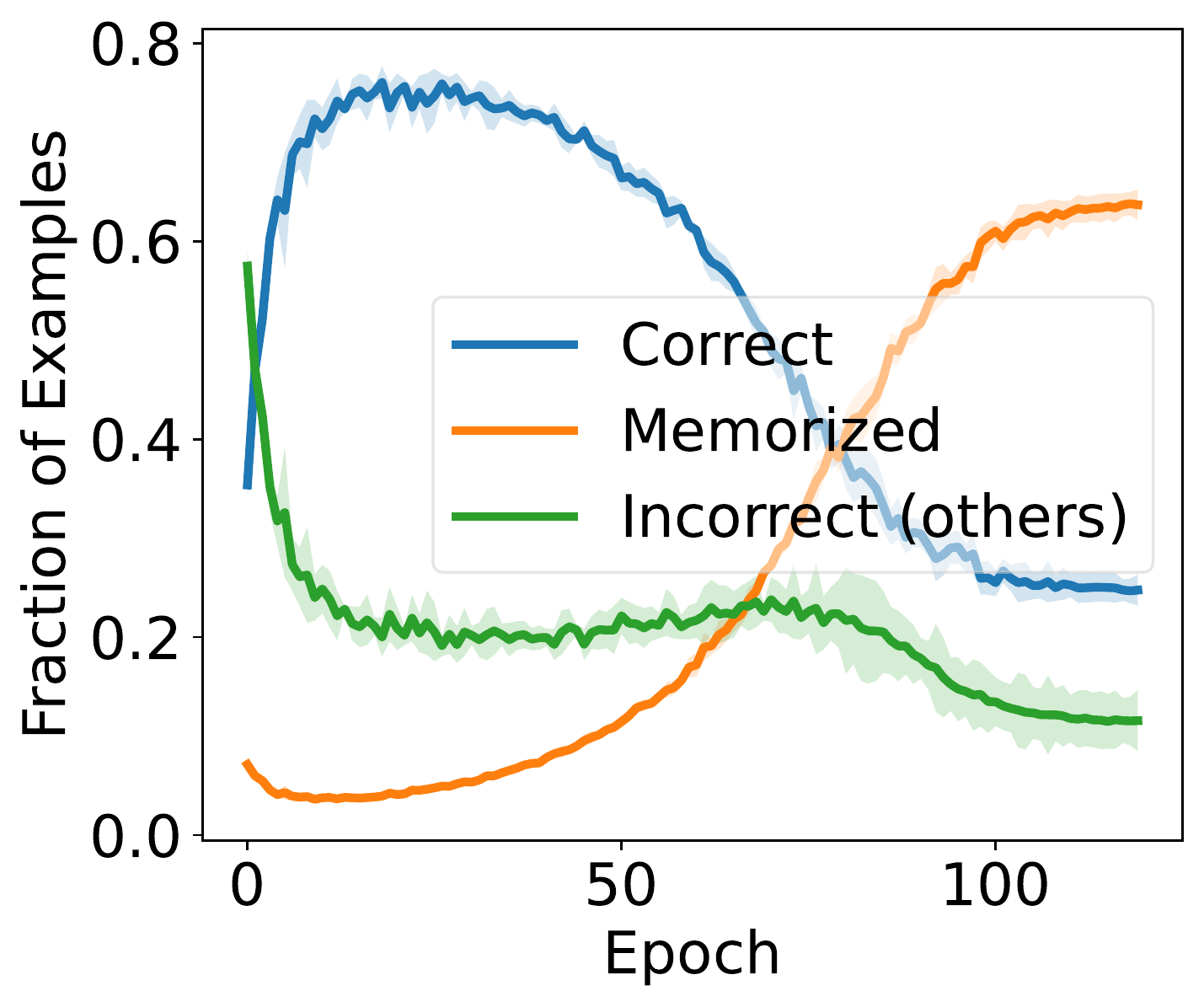}
    \end{subfigure}
    \hfill
    \begin{subfigure}[b]{0.49\linewidth}
        \centering
        \includegraphics[width=\textwidth]{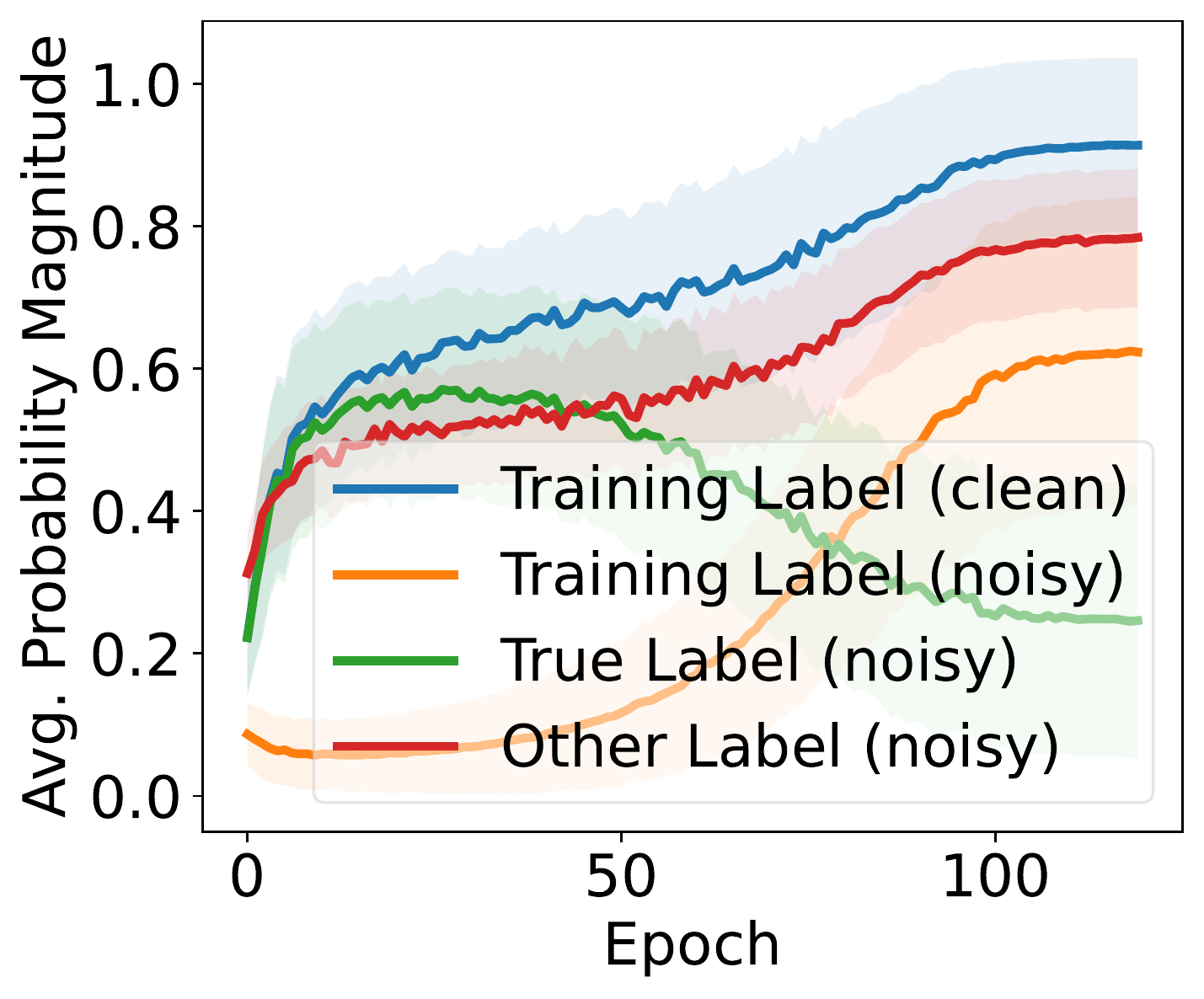}
    \end{subfigure}
    \caption{For ResNet34 models trained with cross-entropy on CIFAR-10 with 25 \% of corrupted instances (averaged over five seeds), the left plot shows the fractions of examples that are correctly classified, whose corrupted training label are memorized, or incorrectly classified with a label other than the ground-truth or training label, confirming the result in \citep{Liu2020EarlyLearningRP}. The right plot illustrates the predicted probability magnitudes for clean or noisy labels.\looseness=-1}
    \label{fig:mem_prob}
\end{figure}

The training dynamics of models have been thoroughly studied in the aforementioned regime 
\citep{DBLP:conf/nips/ChangLM17,DBLP:conf/icml/ArazoOAOM19,Liu2020EarlyLearningRP}, and two learning phases have been identified: Initially, as shown in the left plot of Fig.\ \ref{fig:mem_prob}, the model shows reasonable learning behavior by establishing correct relations between features and targets, classifying even mislabeled instances mostly correctly. In this phase, the loss minimization is dominated by the clean fraction, such that the mislabeling does not affect the learning too much. However, after the clean labels are fit sufficiently well, the learner starts to concentrate predominantly on the mislabeled part, thereby overfitting incorrect labels and harming generalization.

A closer look at the probabilistic predictions in the first learning phase reveals that erroneous training labels can be distinguished from the underlying ground-truth classes by the learner's confidence.  The right plot in Fig.\ \ref{fig:mem_prob} illustrates that models are typically not incentivized to optimize for predicting the corrupted training labels in the first epochs, but infer relatively high probability scores for the (unknown) ground-truth class, at least on average. 
Evidently, the model \textit{itself} could serve as a source for modeling the beliefs about the ground-truth, taking all (mostly clean) instance-label relations into account to reason about the true labels.

This idea has been adopted by label correction methods that predict \emph{pseudo-labels} for instances that appear to be mislabeled, thus suggesting labels that the learner considers more plausible \citep{DBLP:journals/corr/ReedLASER14,DBLP:conf/aaai/WuSX0M21}. While intuitively plausible, a strategy like this is obviously not without risk. Especially in the early phase of the training, replacing a supposedly incorrect label by another one might be too hasty, and may potentially even aggravate the negative effects of label noise.
Instead, as the learner is still in its infancy, it seems advisable to realize a more cautious learning behavior. More concretely, instead of discarding the original training information completely, it might be better to keep it as an option.  

Following this motivation, we propose a \textit{complementary} approach for modeling the learner's belief about the true label: Instead of forcing it to commit to a single label, either the original one or a presumably more plausible alternative, we allow the learner to (re-)label a training instance by a \emph{set} of candidate labels, which may comprise more than a single plausible candidate. More specifically, by retaining the original label and adding other plausible candidates, we deliberately ``ambiguate'' the training information to mitigate the risk of making a mistake in an early phase of the training process. 
To put this idea into practice, we make use of so-called \textit{superset learning} \citep{liu_lo14,osl}, which allows the learner itself to ``disambiguate'' possibly ambiguous training data. 

More precisely, we represent the ambiguous target information in the form of so-called \textit{credal sets}, i.e., sets of probability distributions, to train probabilistic classifiers via generalized risk minimization in the spirit of label relaxation \citep{labelrelaxation}. We realize our approach, which we dub \textit{\pname{}} (\pnames), in an easy off-the-shelf loss function that dynamically derives the target sets from the model predictions without the need of any additional model parameter -- this is implicitly done in the loss calculation, without requiring any change to a conventional learning routine. This way, we combine the simplicity of robust losses with the data modeling capabilities of more complex label correction approaches. We demonstrate the effectiveness of our method on commonly used image classification datasets with both synthetic and real-world noise, confirming the adequacy of our proposed robust loss. 

\section{Related Work}\label{sec:related_work}

Coping with label noise in machine learning is a broad field with an extensive amount of recent literature. Here, we distinguish four views on this issue, namely, robust loss functions, regularization, sample selection, and label correction methods. For a more comprehensive overview, we refer to recent surveys by \citet{Song2020LearningFN} and \citet{cifarn}.

\paragraph{Robust Losses.} The task of designing robust optimization criteria has a long-standing history in classical statistics, e.g., to alleviate the sensitivity towards outliers. As a prominent member of such methods, the mean absolute error (MAE) steps up to mitigate the shortcomings of the mean squared error. When relating to the context of probabilistic classification, robustness of loss functions towards label noise is linked with the symmetry of the function \citep{DBLP:conf/aaai/GhoshKS17}, leading to adaptations of losses such as cross-entropy \citep{Wang2019SymmetricCE,Ma2020NormalizedLF}. A large strain of research proposes to balance MAE and cross-entropy, e.g., by the negative Box-Cox transformation \citep{Zhang2018GeneralizedCE}, or controlling the order of Taylor series for the categorical cross-entropy \citep{Feng2020CanCE}, whereas also alternative loss formulations have been considered \citep{Wei2020WhenOF}. Besides, methodologies accompanying classical losses for robustness have been proposed, such as gradient-clipping \citep{Menon2020CanGC} or sub-gradient optimization \citep{Ma2022BlessingON}. 

\paragraph{Regularization.} Regularizing losses for classification has also been considered as a means to cope with label noise. As one prominent example, label smoothing (LS) \citep{szegedy_ls} has shown similar beneficial properties as loss correction when dealing with label noise \citep{Lukasik2020DoesLS}. Advancements also enable applicability in high noise regimes \citep{Wei2021ToSO}.
Among the first works building upon this observation, \citet{Arpit2017ACL} characterize two phases in learning from data with label noise. First, the model learns to correctly classify most of the instances (including the ground-truth labels of misclassified training instances), followed by the memorization of mislabels. The works shows that explicit regularization, for instance Dropout \cite{dropout}, is effective in combating memorization and improving generalization. Following this, \citet{Liu2020EarlyLearningRP} propose a penalty term to counteract memorization that stems from the (more correct) early learning of the model. Beyond LS, also other forms of ``soft labeling methods'' have been proposed, e.g., as in so-called label distribution learning \citep{DBLP:journals/tip/GaoXXWG17}.
In addition, sparse regularization enforces the model to predict sharp distributions with sparsity \citep{Zhou2021LearningWN}. Lastly, \citet{Iscen2022LearningWN} describe a regularization term based on the consistency of an instance with its neighborhood in feature space.

\paragraph{Sample Selection.} Designed to prevent the aforementioned memorization of mislabeled instances, a wide variety of methods rely on the so-called small loss selection criterion \citep{Jiang2017MentorNetLD,DBLP:conf/ijcai/GuiWT21}. It is intended to distinguish clean samples the model recognizes well, and from which it can learned without distortion. This distinction suggests a range of approaches, including a gradual increase of the clean sample set with increasing model confidence \citep{Shen2018LearningWB,Cheng2021LearningWI,Wang2022ProMixCL}, co-training \citep{Han2018CoteachingRT,Yu2019HowDD,Wei2020CombatingNL}, or re-weighting instances \citep{Chen2021SamplePG}. Furthermore, it makes the complete plethora of classical semi-supervised learning methods amenable to the task of learning from noisy labels. Here, the non-small loss examples are considered as unlabeled in the first place \citep{Li2020DivideMixLW,Nishi2021AugmentationSF,Zheltonozhskii2020ContrastTD}. Often, such methodology is also combined with consistency regularization \citep{consistency_reg}, as prominently used in classical semi-supervised learning \citep{fixmatch,Liu2020EarlyLearningRP,Yao2021JoSRCAC,Liu2022RobustTU}.

\paragraph{Label Correction.} Traditionally, correcting label noise has also been considered by learning probabilistic transition matrices \citep{Goldberger2016TrainingDN,Patrini2016MakingDN,Zhu2022BeyondIL,DBLP:conf/eccv/KyeCYC22}, e.g., to re-weight or correct losses. In this regard, it has been proposed to learn a model and (re-)label based on the model's predictions jointly \citep{DBLP:conf/cvpr/TanakaIYA18}. Closely related to sample selection as discussed before, \citet{DBLP:conf/icml/SongK019} propose to first select clean samples in a co-teaching method, which then refurbishes noisy samples by correcting their label information. Furthermore, \citet{DBLP:conf/icml/ArazoOAOM19} suggest to use two-component mixture models for detecting mislabels that are to be corrected, thus serving as a clean sample criterion. \citet{DBLP:conf/aaai/WuSX0M21} and \citet{DBLP:journals/corr/abs-2302-06810} approach the problem of label correction by a meta-learning technique in the form of a two-stage optimization process of the model and the training labels. In addition, \citet{Liu2022RobustTU} describe an approach to learn the individual instances' noise by an additional over-parameterization. Finally, ensembling has also been considered as a source for label correction \citep{DBLP:conf/ijcai/LuH22}.

\section{Method}\label{sec:method}

The following section outlines the motivation behind our method \pnames{} to model targets in an ambiguous manner, introduces the theoretical foundation, and elaborates on how this approach realizes robustness against label noise.

\subsection{Motivation}\label{sec:method:motivation}

Deep neural network models, often overparameterized with significantly more parameters than required to fit training data points at hand, have become the de-facto standard choice for most practically relevant problems in deep learning. Here, we consider probabilistic models of the form $\hatp : \cX \fromto \P(\cY)$ with $\cX \subset \mathbb{R}^d$ and $\cY=\{y_1, \dots, y_K\}$ being the feature and target space, respectively, and $\P(\cY)$ the class of probability distributions on $\cY$. We denote the training set of size $N$ as $\mathcal{D}_N = \{ (\vec{x}_i,y_i) \}_{i=1}^N \subset \cX \times \cY$, where each instance $\vec{x}_i \in \cX$ is associated with an underlying (true) label $y_i^*$. The latter is conceived as being produced by the underlying (stochastic) data-generating process, i.e., as the realization of a random variable $Y \sim p^*(\cdot \given \vec{x})$.\footnote{In applications such as image classification, the ground-truth conditional distributions $p^*(\cdot \given \vec{x}) \in \mathbb{P}(\cY)$ are often degenerate, i.e., $p^*(y^* \given \vec{x}) \approx 1$ and $p^*(y \given \vec{x}) \approx 0$ for $y \neq y^*$.}
However, the actual label $y_i$ encountered in the training data might be corrupted, which means that $y_i \neq y_i^*$. Reasons for corruption can be manifold, including erroneous measurement and annotation by a human labeler. 

As real-world data typically comprises noise in the annotation process, e.g., due to human labeling errors, their robustness towards label noise is of paramount importance. Alas, models in the regime of deep learning trained with conventional (probabilistic) losses, such as cross-entropy, lack this robustness.
As thoroughly discussed in the analysis of the training dynamics of such models \citep{DBLP:conf/nips/ChangLM17,DBLP:conf/icml/ArazoOAOM19,Liu2020EarlyLearningRP},
two phases can be distinguished in the training of alike models, namely a ``correct concept learning'' and a \emph{memorization} \citep{Arpit2017ACL} or \emph{forgetting} \citep{DBLP:conf/iclr/TonevaSCTBG19} phase (cf. Fig.\ \ref{fig:mem_prob}). While the model relates instances to labels in a meaningful way in the first phase, leading to correct predictions on most training instances, the transition to a more and more memorizing model harms the generalization performance.

\begin{figure}
    \centering
    \includegraphics[width=\linewidth]{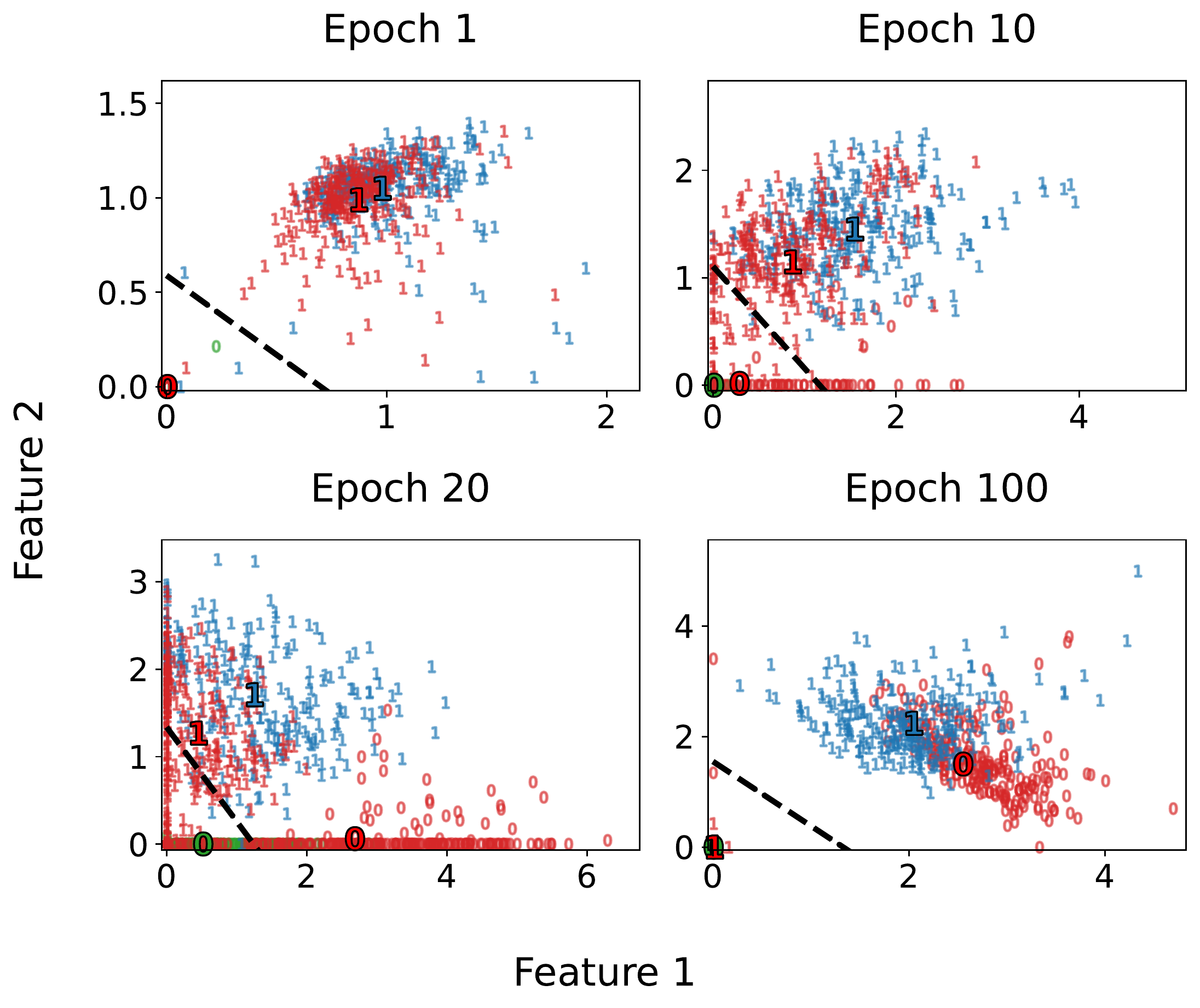}
    \caption{Learned feature representations of the training instances observed at the penultimate layer a MLP comprising an encoder and a classification head at different stages in the training. The data consists of correctly (blue or green resp.) and incorrectly (red) labeled images of zeros and ones from MNIST. The dashed line depicts the linear classifier.}
    \label{fig:learning_dynamics}
\end{figure}
Looking closer at the learning dynamics in an idealized setting, one can observe that overparameterized models project instances of the same ground-truth class $y^*$ to a similar feature embedding, regardless of having observed a correct or corrupted training label in the first training phase.
Fig.\ \ref{fig:learning_dynamics} illustrates the learned feature activations of the penultimate layer of a multi-layer perceptron with a classification head trained on MNIST over the course of the training (see the appendix for further experimental details). In the beginning, the learner predicts a relatively high probability $\hatp(y_i^* \given \vec{x}_i)$ even for noisy instances $(\vec{x}_i, y_i)$, 
as the loss is dominated by the cross-entropy of the clean instances, which leads to a similar marginalization behavior as in the non-corrupted case. Here, the proximity of mislabeled instances to the decision boundary can be related to the ``hardness'' of instances to be learned, as also studied by \citet{DBLP:conf/nips/ChangLM17}.
In later stages, the feature activations of the noisy instances shift in a rotating manner, successively being pushed towards the other discriminatory face of the decision boundary. This goes hand in hand with a decrease and eventual increase of the predicted probability scores $\max_{y \in \cY} \hatp(y \given \vec{x})$, consistent with the observations made in Fig.\ \ref{fig:mem_prob}.

It appears natural to seek for means that keep corrupted instances $(\cdot, y_i)$ close to clean samples $(\cdot, y_i^*)$ in the feature space, and not let the learning bias in the later stages pull them over towards the label corruption.
In the context of the overall optimization, the model itself represents a source of justification for considering a class label as a plausible outcome. Hence, we suggest to use the model predictions \textit{simultaneously} with the training labels for modeling our beliefs about the ground-truth. Consequently, we shall consult not only the individual training labels, but also the \textit{complete} training dataset that found its way into the current model in conjunction, complementing the former as a second piece of evidence. 
This represents a distillation of knowledge obtained from the (mostly clean) data at hand, and we argue that the confidence $\hatp$ is a suitable surrogate to represent plausibility in this regard.

\subsection{Credal Labeling}\label{sec:method_credal_labeling}

Inquiring the model prediction $\hatp(\vec{x}_i)$ in addition to the training label $y_i$ may call for augmenting the (hitherto single) label by an additional plausible candidate label $\hat{y}_i \in \argmax_{y' \in \cY} \hatp(y' \given \vec{x}_i)$ with $\hat{y}_i \neq y_i$, making the target effectively ambiguous and hence less convenient for conventional point-wise classification losses. 
However, from a data modeling perspective, it is important to recognize that the imprecisiation of the target information pays off with higher validity, i.e., it is more likely that the true label is covered by the target labels. This is completely in line with the idea of \textit{data imprecisiation} in the context of so-called \textit{superset learning} \citep{DBLP:journals/ijar/LienenH21}. Thus, ambiguating presumably corrupt training information appears to be a useful means to counter the influence of mislabeling.

Before detailing the representation of the aforementioned beliefs, we shall revisit the conventional probabilistic learning setting. To train probabilistic classifiers $\hatp$ as specified above, e.g., by minimizing the cross-entropy loss, traditional methods transform deterministic target labels $y_i \in \cY$ as commonly observed in classification datasets into degenerate probability distributions $p_{y_i} \in \P(\cY)$ with $p_{y_i}(y_i)=1$ and $p_{y_i}(y)=0$ for $y \neq y_i$. As a result, the predicted distribution $\hatp$ can be compared to $p_{y_i}$ using a probabilistic loss $\mathcal{L} : \P(\cY) \times \P(\cY) \fromto \R$ to be minimized. It is easy to see that $p_{y_i}$ assigns full plausibility to the observed training label $y_i$, while the other labels are regarded as fully implausible.

Ambiguity in a probabilistic sense can be attained by enriching the representation  of the target distribution in a set-valued way. To this end, \citet{labelrelaxation} propose to use \textit{credal sets}, i.e., sets of probability distributions, as a means to express beliefs about the true class conditional distribution $p^*$. This allows one, for example, to not only consider $p_{y_i}$ for the training label $y_i$ as a fully plausible target distribution, but also $p_{y}$ for a plausible candidate label $y \neq y_i$ -- and interpolations between these extremes, i.e., distributions $p = \lambda p_{y_i} + (1-\lambda) p_{y}$ assigning probability $\lambda$ to label $y_i$ and $1-\lambda$ to $y$. More generally, a credal set can be any (convex) set of probability distributions, including, for example, ``relaxations'' of extreme distributions $p_{y_i}$, which reduce the probability for $y_i$ to $1- \epsilon < 1$ and reassign the remaining mass of $\epsilon > 0$ to the other labels. Such distributions appear to be meaningful candidates as target distributions, because a degenerate ground truth $p^* = p_{y_i}$ is actually not very likely \citep{labelrelaxation}. 

Possibility theory \citep{Dubois2004PossibilityTP} offers a compact representation of credal sets in the form of \emph{upper bounds} on probabilities, encoded as \textit{possibility distributions} $\pi : \mathcal{Y} \fromto [0,1]$. Intuitively, for each outcome $y \in \cY$, $\pi(y)$ defines an upper bound on the true probability $p^*(y)$. More generally, a distribution $\pi$ induces a \emph{possibility measure} $\Pi(Y)=\max_{y\in Y} \pi(y)$ on $\cY$, which in turn defines an upper bound on any event $Y \subset \cY$. In our case, a distribution $\pi_i$ associated with an instance $\vec{x}_i$ assigns a possibility (or plausibility) to each class in $\cY$ for being the true outcome associated with $\vec{x}_i$, and $\pi_i(y)$ can be interpreted as an upper bound on $p^*(y \given \vec{x}_i)$. Thus, the credal set of distributions considered as candidates for $p^*(y \given \vec{x}_i)$ is given as follows:
\begin{equation}\label{eq:credalset}
Q_{\pi_i} \defeq \Big\{ p \in \mathbb{P}(\mathcal{Y}) \, \vert \, \forall \, Y \subseteq \mathcal{Y} : \, \sum_{y \in Y} p(y) \leq \max_{y \in Y} \pi_i(y) \Big\}
\end{equation}

\subsection{Data Ambiguation for Robust Learning}\label{sec:method:confidence_based_elication}

With the above approach to modeling ambiguous targets in a probabilistic manner, we can put the idea of a robust ambiguation method on a rigorous theoretic foundation. As a concrete model of the target label for a training instance $(\vec{x}_i, y_i)$, we propose the following confidence-thresholded possibility distribution $\pi_i$:
	\begin{equation} \label{eq:ambig_label}
    \pi_i(y) = \begin{cases}
    1 & \text{if }  y = y_i \vee \hatp(y \given \vec{x}_i) \geq \beta \\
    \alpha & \text{otherwise }
\end{cases}  \enspace ,
\end{equation}
where $\beta \in [0,1]$ denotes the confidence threshold for the prediction $\hatp(y \, | \, \vec{x}_i)$, and $\alpha \in [0,1)$ is the label relaxation parameter \citep{labelrelaxation}. Thus, while the original label $y_i$ remains fully plausible ($\pi_i(y_i)=1$), it might be complemented by other candidates: as soon as the predicted probability for another label exceeds $\beta$, it is deemed a fully plausible candidate, too. Besides, by assigning a small degree of possibility $\alpha$ to all remaining classes, these are not completely excluded either. Both $\alpha$ and $\beta$ are treated as hyperparameters.

To learn from credal targets $\mathcal{Q}_{\pi_i}$ induced by $\pi_i$, i.e., from the ambiguated training data $\{ (\vec{x}_i, \mathcal{Q}_{\pi_i}) \}$, we make use of a method for superset learning that is based on minimizing the \textit{optimistic superset loss} \citep{supsetlearning}. The latter generalizes a standard loss $\mathcal{L}$ for ``point predictions'' as follows:\looseness=-1
\begin{equation}\label{eq:gloss}
\mathcal{L}^*(\mathcal{Q}_{\pi}, \hatp ) \defeq \min_{p \in \mathcal{Q}_{\pi}} \mathcal{L}(p, \hatp) \, .
\end{equation}
As argued by \citet{supsetlearning}, minimizing OSL on the ambiguated training data is suitable for inducing a model while simultaneously disambiguating the data. In other words, the learner makes a most favourable selection among the candidate labels, and simultaneously fits a model to that selection (by minimizing the original loss $\mathcal{L}$).  

A common choice for $\mathcal{L}$ is the Kullback-Leibler divergence $D_{KL}$, for which (\ref{eq:gloss}) simplifies to
\begin{equation}
    \mathcal{L}^*(\mathcal{Q}_{\pi} , \hatp) = \begin{cases}
        0  & \text{if } \hatp \in \mathcal{Q}_{\pi} \\
        D_{KL}(p^r \, || \, \hatp) & \text{otherwise} 
    \end{cases} \, ,
    \label{eq:lstar_kldiv}
\end{equation}
where 
\begin{equation*}
    p^r(y) = \begin{cases}
(1 - \alpha) \cdot \frac{ \hatp(y)}{\sum_{y' \in \mathcal{Y} : \pi_i(y')=1} \hatp(y')} & \text{if } \pi_i(y) = 1 \\
\alpha \cdot \frac{ \hatp(y)}{\sum_{y' \in \mathcal{Y} : \pi_i(y')=\alpha} \hatp(y')} & \text{otherwise}
\end{cases}
\label{eq:targetvec}
\end{equation*}
is projecting $\hatp$ onto the boundary of $\mathcal{Q}_{\pi}$. This loss is provably convex and has the same computational complexity as standard losses such as cross-entropy \cite{labelrelaxation}. 

\begin{figure}[t]
    \centering
    \includegraphics[width=\linewidth]{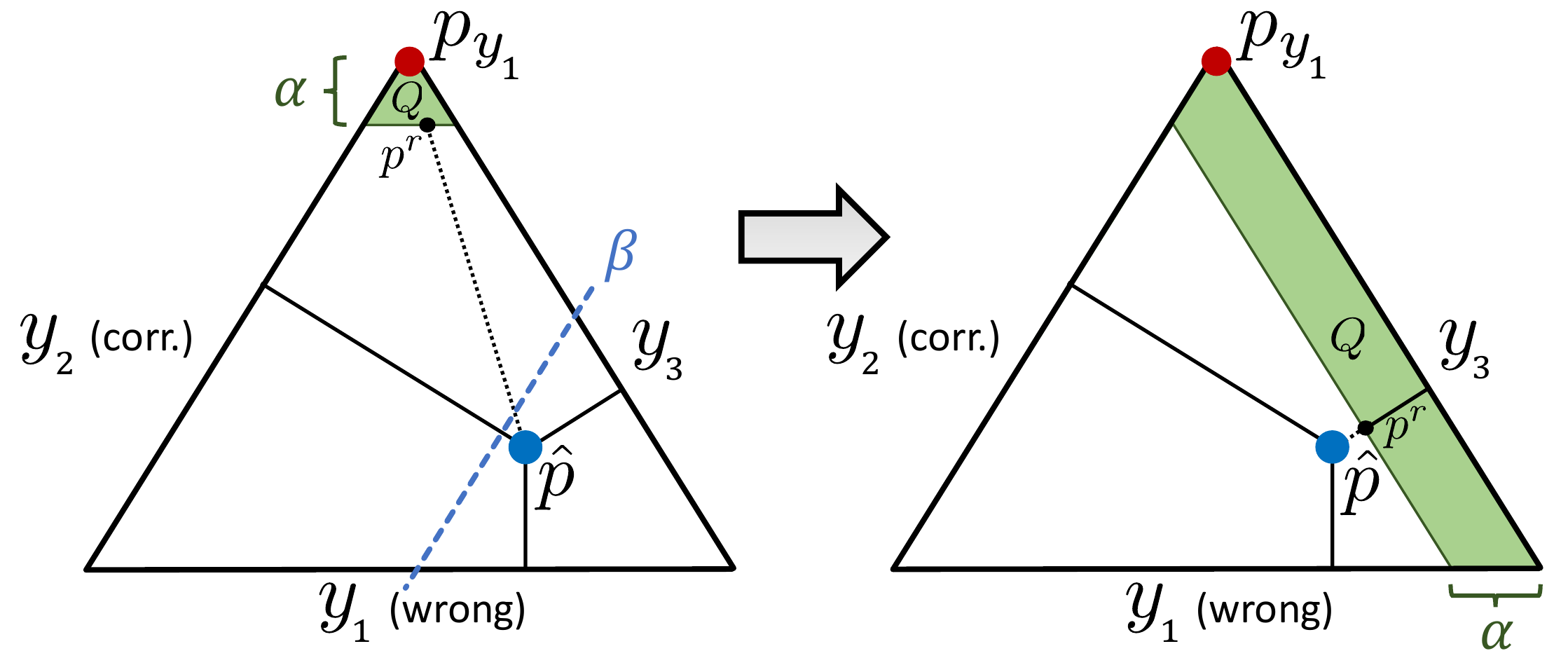}
    \caption{A barycentric visualization of the confidence-thresholded ambiguation for a corrupt training label $y_1$ and a ground-truth $y_2$ in the target space $\cY = \{y_1, y_2, y_3\}$: Starting from a credal set $\mathcal{Q}$ centered at $p_{y_1}$ (left plot), the prediction $\hatp$ predicts a probability mass greater than $\beta$ for $y_2$. Consequently, full possibility is assigned to $y_2$, leading to $\mathcal{Q}$ as shown to the right.}
    \label{fig:barycentric}
\end{figure}

\begin{algorithm}[t]
    \caption{Robust Data Ambiguation (RDA) Loss}
    \label{alg:rda}
    \begin{algorithmic}[1]
        \REQUIRE Training instance $(\vec{x}, y) \in \cX \times \cY$, model prediction $\hatp(\vec{x}) \in \P(\cY)$, confidence threshold $\beta \in [0,1]$, relaxation parameter $\alpha \in [0,1)$
        \STATE Construct $\pi$ as in Eq.\ (\ref{eq:lstar_kldiv}) with \begin{equation*}
            \pi(y') = \begin{cases}
            1 & \text{if }  y' = y \vee \hatp(y' \given \vec{x}) \geq \beta \\
            \alpha & \text{otherwise }
            \end{cases}
            \end{equation*}
        \RETURN $\mathcal{L}^*(Q_\pi, \hatp(\vec{x}))$ as specified in Eq.\ (\ref{eq:lstar_kldiv}), where $Q_\pi$ is derived from $\pi$
    \end{algorithmic}
\end{algorithm}

Fig.\ \ref{fig:barycentric} illustrates the core idea of our method, which we will refer to as \textit{\pname} (\pnames). Imagine an instance $\vec{x}$ with true label $y_2$ but corrupted training information $y_1$. Initially, we observe a probabilistic target centered at $p_{y_1}$ (and potentially  relaxed by some $\alpha > 0$). Without changing the target, label relaxation would compute the loss by comparing the prediction $\hatp$ to a (precise) distribution $p^r$ projecting $\hatp$ towards $p_{y_1}$. In this example, the model predicts $y_2$ with a confidence exceeding $\beta$. With our method, full plausibility would be assigned to $y_2$ in addition to $y_1$, leading to a credal set $\mathcal{Q}$ as shown in the right plot. 
To compute the loss, $\hatp$ is now compared to a less extreme target $p^r$ that is projected onto the larger credal set, no longer urging the learner to predict distributions close to $p_{y_1}$. A more technical description of RDA is provided in Alg.\ \ref{alg:rda}, confirming the simplicity of our proposal.\looseness=-1

As can be seen, by ambiguating the target set, we relieve the learner from memorizing wrong training labels. At the same time, we remain cautious and try to avoid committing to potentially erroneous predictions. Thereby, learning from noisy labels is de-emphasized while the optimization on clean samples remains unaffected. At the same time, driven by the loss minimization on similar instances that are correctly labeled with $y_2$, the imprecisiation allows the predictions $\hatp \in \mathcal{Q}$ to evolve towards the true target $p_{y_2}$.

Surpassing the parameter $\beta$ can be seen as an indication for mislabeling, i.e., the prediction $\hatp$ suggests that a different label $y \neq y_i$ is a plausible candidate for representing the true outcome $y^*$. As such, $\beta$ can be used to control the ``cautiousness'' in suspecting mislabeling: High values for $\beta$ force that only highly confident predictions shall adjust the target, whereas less extreme values result in a more eager addition of candidates. Conversely, since classes $y$ with $\hatp(y) < \beta$ are considered as implausible, the threshold $\beta$ could also be interpreted as a criterion for exclusion of candidates. As shown in a more extensive analysis of the $\beta$ parameter, the former interpretation is practically more useful in the case of robust classification.

Generally, any value in $[0,1]$ could be chosen for $\beta$, perhaps also depending on the training progress. As a simple yet effective rule of thumb, we suggest to vary $\beta$ over the course of time: As the model is relatively uncertain in the first epochs, one should not spent too much attention to the  predictions. However, with further progress, $\hatp$ becomes more informative for the confidence-driven possibility elicitation, suggesting to use smaller $\beta$ values. Empirically, we found the following decay function to yield good performance:
\begin{equation}\label{eq:cosine_decay}
    \beta_T = \beta_1 + \dfrac{1}{2} \left( \beta_0 - \beta_1 \right)\left(1 + \cos\left(\dfrac{T}{T_{max}}\pi \right) \right) \enspace ,
\end{equation}
where $T$ and $T_{max}$ denote the current and maximum number of epochs, respectively, while $\beta_0$ and $\beta_1$ represent the start and end values for $\beta$. Nevertheless, as will be shown in the empirical evaluation, also static values for $\beta$ work reasonably well, such that the number of additional hyperparameters can be reduced to a single one.

\section{Experiments}\label{sec:experiments}

To demonstrate our method's effectiveness in coping with label noise, we conduct an empirical analysis on several image classification datasets as a practically relevant problem domain, although \pnames{} is not specifically tailored to this domain and does not use any modality-specific components. Here, we consider CIFAR-10/-100 \citep{CIFAR} as well as the large-scale datasets WebVision \citep{webvision} and Clothing1M \citep{clothing1m} as benchmarks. For the former, we model synthetic noise by both symmetrically and asymmetrically randomizing the labels for a fraction of examples, while WebVision and Clothing1M comprise real-world noise by their underlying crawling process. Moreover, we report results on CIFAR-10(0)N \citep{cifarn} as another real-world noise dataset based on human annotators.
We refer to the appendix for a more detailed description of the datasets and the corruption process, as well as experiments on additional benchmark datasets.\looseness=-1

As baselines, we take a wide range of commonly applied loss functions into account. Proceeding from the conventional cross-entropy (CE), we report results for the regularized CE adaptations label smoothing (LS) and label relaxation (LR), as well as the popular robust loss functions generalized cross-entropy (GCE) \citep{Zhang2018GeneralizedCE}, normalized cross-entropy (NCE) \citep{Ma2020NormalizedLF}, combinations of NCE with AUL and AGCE \citep{DBLP:conf/icml/ZhouLJGJ21} and CORES \cite{Cheng2021LearningWI}. For completeness, we also report results for ELR \citep{Liu2020EarlyLearningRP} and SOP \citep{Liu2022RobustTU} as two state-of-the-art representatives of regularization and label correction methods, respectively. While all other losses can be used off-the-shelf, these two have additional parameters (to track the label noise per instance), giving them an arguably unfair advantage. In the appendix, we show further results for a natural baseline to our approach in the realm of superset learning, as well as experiments that combine our method with sample selection.

We follow common practice and evaluate methods against label noise in the regime of overparameterized models by training ResNet34 models on the smaller scale datasets CIFAR-10/-100. For the larger datasets, we consider ResNet50 models pretrained on ImageNet. All models use the same training procedure and optimizer. A more thorough overview of experimental details, such as hyperparameters and their optimization, as well as the technical infrastructure, can be found in the appendix. We repeated each run five times with different random seeds, reporting the accuracy on the test splits to quantify generalization.

\subsection{Synthetic Noise}\label{sec:experiments:synthetic}

\begin{table*}[ht]
    \centering
    \fontsize{9pt}{9pt}\selectfont
    \begin{tabular}{c|c|ccc|ccc}
        \toprule
        \multirow{3}{*}{Loss} & \multirowcell{3}{Add.\\Param.} & \multicolumn{3}{c|}{CIFAR-10} & \multicolumn{3}{c}{CIFAR-100} \\
        & & \multicolumn{3}{c|}{Sym.} & \multicolumn{3}{c}{Sym.} \\
        & & 25 \% & 50 \% & 75 \% & 25 \% & 50 \% & 75 \% \\
        \midrule
        CE & \xmark & 79.05 {\small $\pm${}0.67} & 55.03 {\small $\pm${}1.02} & 30.03 {\small $\pm${}0.74} & 58.27 {\small $\pm${}0.36} & 37.16 {\small $\pm${}0.46} & 13.66 {\small $\pm${}0.45}  \\
        LS ($\alpha=0.1$) & \xmark & 76.66 {\small $\pm${}0.69} & 53.95 {\small $\pm${}1.47} & 29.03 {\small $\pm${}1.21} & 59.75 {\small $\pm${}0.24} & 37.61 {\small $\pm${}0.61} & 13.53 {\small $\pm${}0.51}  \\
        LS ($\alpha=0.25$) & \xmark & 77.48 {\small $\pm${}0.32} & 53.08 {\small $\pm${}1.95} & 28.29 {\small $\pm${}0.65} & 59.84 {\small $\pm${}0.57} & 39.80 {\small $\pm${}0.38} & 14.18 {\small $\pm${}0.44}  \\
        LR ($\alpha=0.1$) & \xmark & 80.53 {\small $\pm${}0.39} & 57.55 {\small $\pm${}0.95} & 29.83 {\small $\pm${}0.87} & 57.52 {\small $\pm${}0.58} & 36.77 {\small $\pm${}0.54} & 13.23 {\small $\pm${}0.14}  \\
        LR ($\alpha=0.25$) & \xmark & 80.43 {\small $\pm${}0.09} & 60.18 {\small $\pm${}1.01} & 31.36 {\small $\pm${}0.91} &  57.67 {\small $\pm${}0.11} & 37.15 {\small $\pm${}0.14} & 13.41 {\small $\pm${}0.24}  \\
        \midrule
        GCE & \xmark & 90.82 {\small $\pm${}0.10} & 83.36 {\small $\pm${}0.65} & 54.34 {\small $\pm${}0.37} & 68.06 {\small $\pm${}0.31} & 58.66 {\small $\pm${}0.28} & \textbf{26.85} {\small $\pm${}1.28}  \\
        NCE & \xmark & 79.05 {\small $\pm${}0.12} & 63.94 {\small $\pm${}1.74} & 38.23 {\small $\pm${}2.63} & 19.32 {\small $\pm${}0.81} & 11.09 {\small $\pm${}1.03} & 6.12 {\small $\pm${}7.57}  \\
        NCE+AGCE & \xmark & 87.57 {\small $\pm${}0.10} & 83.05 {\small $\pm${}0.81} & 51.16 {\small $\pm${}6.44} & 64.15 {\small $\pm${}0.23} & 39.64 {\small $\pm${}1.66} & 7.67 {\small $\pm${}1.25}  \\
        NCE+AUL & \xmark & 88.89 {\small $\pm${}0.29} & 84.18 {\small $\pm${}0.42} & \textbf{65.98} {\small $\pm${}1.56}  & 69.76 {\small $\pm${}0.31} & 57.41 {\small $\pm${}0.41} & 17.72 {\small $\pm${}1.27} \\
        CORES & \xmark & 88.60 {\small $\pm${}0.28} & 82.44 {\small $\pm${}0.29} & 47.32 {\small $\pm${}17.03} & 60.36 {\small $\pm${}0.67} & 46.01 {\small $\pm${}0.44} & 18.23 {\small $\pm${}0.28} \\
         \midrule
         \pnames{} (ours) & \xmark & \textbf{91.48} {\small $\pm${}0.22} & \textbf{86.47} {\small $\pm${}0.42} & 48.11 {\small $\pm${}15.41} & \textbf{70.03} {\small $\pm${}0.32} & \textbf{59.83} {\small $\pm${}1.15} & 26.75 {\small $\pm${}8.83} \\
         \midrule \midrule
         ELR & \cmark & 92.45 {\small $\pm${}0.08} & 88.39 {\small $\pm${}0.36} & 72.58 {\small $\pm${}1.63} & \underline{73.66} {\small $\pm${}1.87} & 48.72 {\small $\pm${}26.93} & 38.35 {\small $\pm${}10.26}  \\
         SOP & \cmark & \underline{92.58} {\small $\pm${}0.08} & \underline{89.21} {\small $\pm${}0.33} & \underline{76.16} {\small $\pm${}4.88} & 72.04 {\small $\pm${}0.67} & \underline{64.28} {\small $\pm${}1.44} & \underline{40.59} {\small $\pm${}1.62} \\
         \bottomrule
    \end{tabular}
    \caption{Test accuracies and standard deviations on the test split for models trained on CIFAR-10(0) with (symmetric) synthetic noise. The results are averaged over runs with different seeds, bold entries mark the best method without any additional model parameters. Underlined results indicate the best method overall.}
    \label{tab:syn}
\end{table*}

Table\ \ref{tab:syn} reports the results for the synthetic corruptions on CIFAR-10/-100. As can be seen, our approach provides consistent improvements in terms of generalization performance over the robust off-the-shelf loss functions. Cross-entropy and its regularized adaptations LS and LR appear sensitive to label noise, which confirms the need for loss robustness. Interestingly, although being slightly inferior in most cases of symmetric noise, our method appears still competitive compared to ELR and SOP, despite their increased expressivity through additional parameters. For asymmetric noise, as presented in the appendix, our method could even outperform such methods. Nevertheless, it becomes less effective with an increasing level of noise. 

\begin{figure}[t]
    \centering
    \includegraphics[width=0.8\columnwidth]{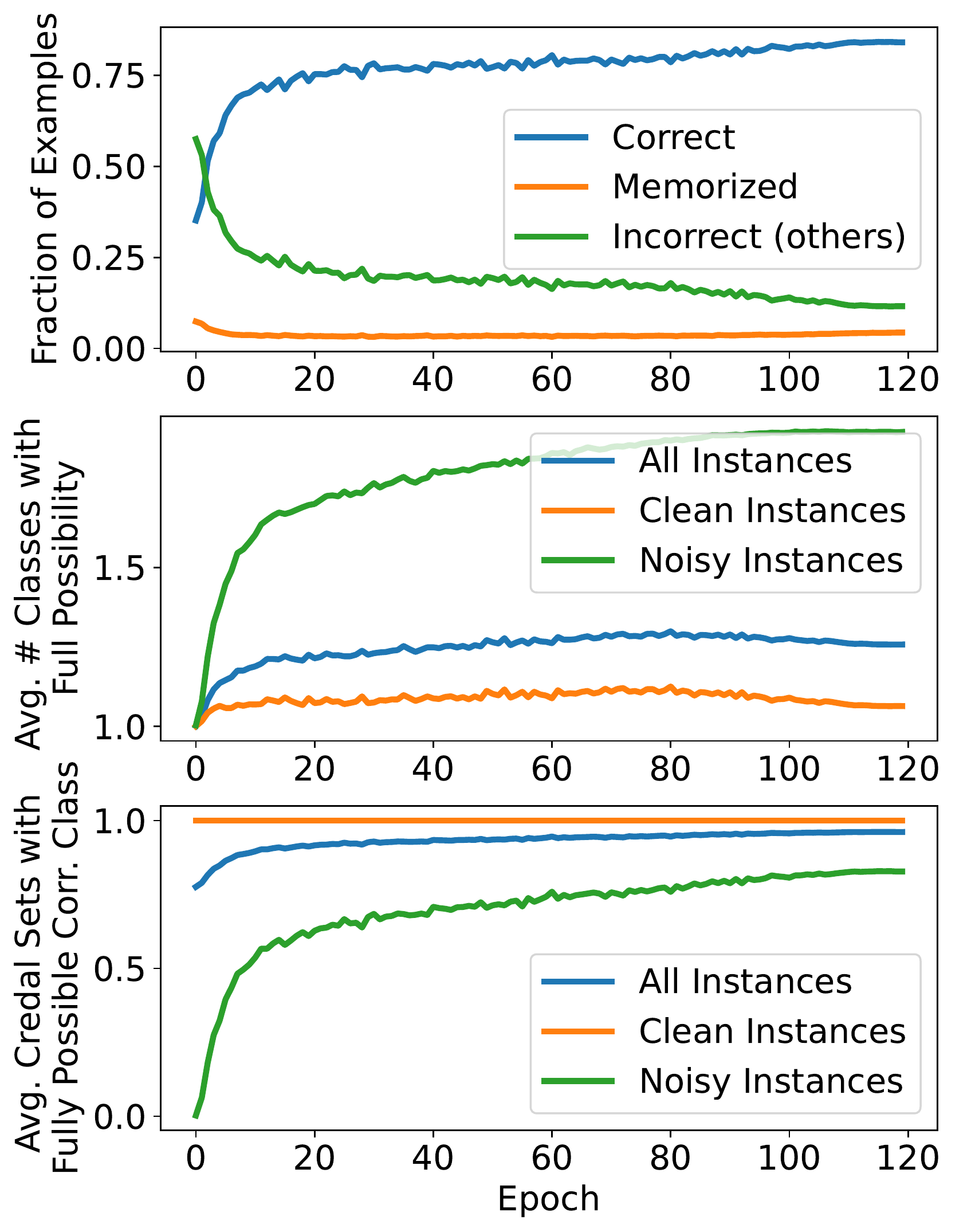}
    \caption{The top plot shows the fraction of mislabeled training instances for which the models predict the ground-truth (blue), the wrong training label (orange) or a different label (green). The middle and bottom plots show the credal set size and validity respectively. All plots are averaged over the five runs on CIFAR-10 with 50 \% synthetic symmetric noise.\looseness=-1}
    \label{fig:syn}
\end{figure}

When looking at the learning dynamics of training with RDA, Fig.\ \ref{fig:syn} reveals an effective ambiguation in the course of the learning process. The left plot shows the attenuation of any memorization while improving the correctness of model predictions for the wrongly labeled instances at the same time. The plot in the middle depicts an increase of the credal set size in terms of classes with full plausibility for the noisy instances. Together with the right plot showing the validity of the credal sets, i.e., the fraction of credal sets that assign full possibility to the ground-truth class, one can easily see that the ambiguation is indeed able to select the ground-truth class as training label. Furthermore, the credal set size for clean instances is barely affected, which again confirms the adequacy of our model. 
Notably, our method also shows self-correcting behavior after ambiguating with a wrong class midway. While the validity of the credal sets increases (roughly) monotonically, the set sizes become smaller towards the end of the training.
In the appendix, we provide additional plots in other noise settings showing consistent effects.

\subsection{Real-World Noise}\label{sec:experiments:realworld}

\begin{table*}[t]
    \centering
    \fontsize{9pt}{9pt}\selectfont
    \begin{tabular}{c|c|ccccc|c}
        \toprule
        \multirow{2}{*}{Loss} & \multirowcell{2}{Add.\\Param.} & \multicolumn{5}{c|}{CIFAR-10N} & \multicolumn{1}{c}{CIFAR-100N} \\
        &  & Random 1 & Random 2 & Random 3 & Aggregate & Worst & Noisy \\
        \midrule
        CE & \xmark  & 82.96 {\small $\pm${}0.23} & 83.16 {\small $\pm${}0.52} & 83.49 {\small $\pm${}0.34} & 88.74 {\small $\pm${}0.13} & 64.93 {\small $\pm${}0.79} & 52.88 {\small $\pm${}0.14} \\
        LS ($\alpha=0.1$) & \xmark & 82.76 {\small $\pm${}0.47} & 82.10 {\small $\pm${}0.21} & 82.12 {\small $\pm${}0.37} & 88.63 {\small $\pm${}0.11} & 63.10 {\small $\pm${}0.38} & 53.48 {\small $\pm${}0.45} \\
        LS ($\alpha=0.25$) & \xmark & 82.95 {\small $\pm${}1.57} & 83.86 {\small $\pm${}2.05} & 82.61 {\small $\pm${}0.25} & 87.03 {\small $\pm${}2.29} & 66.14 {\small $\pm${}6.89} & 53.98 {\small $\pm${}0.27} \\
        LR ($\alpha=0.1$) & \xmark & 83.00 {\small $\pm${}0.36} & 82.64 {\small $\pm${}0.31} & 82.82 {\small $\pm${}0.21} & 88.41 {\small $\pm${}0.29} & 66.62 {\small $\pm${}0.33} & 52.01 {\small $\pm${}0.04} \\
        LR ($\alpha=0.25$) & \xmark & 82.14 {\small $\pm${}0.49} & 81.87 {\small $\pm${}0.34} & 82.46 {\small $\pm${}0.11} & 88.07 {\small $\pm${}0.45} & 66.44 {\small $\pm${}0.14} & 52.22 {\small $\pm${}0.29} \\
        \midrule
        GCE & \xmark & 88.85 {\small $\pm${}0.19} & 88.96 {\small $\pm${}0.32} & 88.73 {\small $\pm${}0.11} & 90.85 {\small $\pm${}0.32} & 77.24 {\small $\pm${}0.47} & 55.43 {\small $\pm${}0.47} \\
        NCE & \xmark & 81.88 {\small $\pm${}0.27} & 81.02 {\small $\pm${}0.32} & 81.48 {\small $\pm${}0.13} & 84.62 {\small $\pm${}0.49} & 69.40 {\small $\pm${}0.10} & 21.12 {\small $\pm${}0.67} \\
        NCE+AGCE & \xmark & 89.48 {\small $\pm${}0.28} & 88.95 {\small $\pm${}0.10} & 89.25 {\small $\pm${}0.29} & 90.65 {\small $\pm${}0.44} & 81.27 {\small $\pm${}0.44} & 51.42 {\small $\pm${}0.65} \\
        NCE+AUL & \xmark & 89.42 {\small $\pm${}0.22} & 89.36 {\small $\pm${}0.15} & 88.94 {\small $\pm${}0.55}  & 90.92 {\small $\pm${}0.19} & 81.28 {\small $\pm${}0.47} & 56.58 {\small $\pm${}0.41} \\
        CORES & \xmark & 86.09 {\small $\pm${}0.57} & 86.48 {\small $\pm${}0.27} & 86.02 {\small $\pm${}0.22} & 89.23 {\small $\pm${}0.10} & 76.80 {\small $\pm${}0.96} &  53.04 {\small $\pm${}0.29} \\
         \midrule
         \pnames{} (ours) & \xmark & \textbf{90.43} {\small $\pm${}0.03} & \textbf{90.09} {\small $\pm${}0.29} & \textbf{90.40} {\small $\pm${}0.01} & \textbf{91.71} {\small $\pm${}0.38} & \textbf{82.91} {\small $\pm${}0.83} & \textbf{59.22} {\small $\pm${}0.26} \\
         \midrule \midrule
         ELR & \cmark & \underline{91.35} {\small $\pm${}0.29} & \underline{91.46} {\small $\pm${}0.29} & \underline{91.39} {\small $\pm${}0.03} & \underline{92.68} {\small $\pm${}0.03} & \underline{84.82} {\small $\pm${}0.42} & \underline{62.80} {\small $\pm${}0.27} \\
         SOP & \cmark & 89.16 {\small $\pm${}0.40} & 89.02 {\small $\pm${}0.33} & 88.99 {\small $\pm${}0.31} & 90.54 {\small $\pm${}0.16} & 80.65 {\small $\pm${}0.13} & 59.32 {\small $\pm${}0.41} \\
         \bottomrule
    \end{tabular}
    \caption{Test accuracies and standard deviations on the test split for models trained on CIFAR-10(0)N with real-world noise, the scores on the clean splits can be found in the appendix. The results are averaged over runs with different seeds.}
    \label{tab:real_world_cifarn}
\end{table*}

In coherence with the previous observations, our robust ambiguation loss also works reasonably well for real-world noise. As presented in Table\ \ref{tab:real_world_cifarn}, \pnames{} leads to superior generalization performance compared to baseline losses without any additional parameters in almost any case. Moreover, it also consistently outperforms SOP on CIFAR-10N, whereas it leads to similar results for CIFAR-100N.

\begin{table}[t]
\centering
    \fontsize{10pt}{10pt}\selectfont
    \begin{tabular}{c|c|c}
        \toprule
         Loss & WebVision & Clothing1M \\
         \midrule
         CE & 66.96 & 68.04 \\
         GCE & 61.76 & 69.75 \\
         AGCE & 69.4 & - \\
         NCE+AGCE & 67.12 & - \\
         \midrule
         \pnames{} (ours) & \textbf{70.23} & \textbf{71.42} \\
         \bottomrule
    \end{tabular}%
    \caption{Large-scale test accuracies on WebVision and Clothing1M using ResNet50 models. The baseline results for the two datasets are taken from \citep{DBLP:conf/icml/ZhouLJGJ21} and \citep{Liu2020EarlyLearningRP}, respectively.\looseness=-1}
    \label{tab:real_world_ls}
\end{table}

For the large-scale datasets WebVision and Clothing1M, for which results are presented in Table\ \ref{tab:real_world_ls}, the differences between the baselines and our approach appear to be rather modest but are still in favor of our method.

\section{Conclusion}

Large models are typically prone to memorizing noisy labels in classification tasks. To address this issue, various loss functions have been proposed that enhance the robustness of conventional loss functions against label noise. Although such techniques are appealing due to their simplicity, they typically lack the capacity to incorporate additional knowledge regarding the instances, such as beliefs about the true label. In response, pseudo-labeling methods for label correction have emerged. They allow for modeling target information in a more sophisticated way, albeit at the cost of an increased training complexity.

To address the shortcomings of previous methods, our approach advocates the idea of weakening (ambiguating) the training information. In a sense, it unifies the two directions pursued so far, namely, modifying the learning procedure (loss function) and modifying the data (selection, correction). By allowing the learner to (re-)label training data with credal sets of probabilistic labels, the approach becomes very flexible. For the specific type of label ambiguation we proposed in this paper, we could show that learning from the re-labeled data is equivalent to minimizing a ``robustified'' loss function that is easy to compute. Our empirical evaluation confirms the adequacy of our proposal.

Our approach suggests several directions for future research. For example, a more informed choice of $\beta$ and $\alpha$ could be realized by quantifying the epistemic uncertainty \citep{DBLP:journals/ml/HullermeierW21} of individual predictions, as highly uncertain guesses should be considered with caution. Also, our method could be leveraged for various downstream tasks, e.g., to detect anomalies in mostly homogeneous data. Finally, as we are currently only considering the predictions at a time, further improvements could be achieved by taking the dynamics of the training into account.

\section*{Acknowledgements}

This work was partially supported by the German Research Foundation (DFG) within the Collaborative Research Center ``On-The-Fly Computing'' (CRC 901 project no.~160364472). Moreover, the authors gratefully acknowledge the funding of this project by computing time provided by the Paderborn Center for Parallel Computing (PC$^2$).

\bibliography{main}

\newpage


\section*{Supplementary material}

# Mitigating Label Noise through Data Ambiguation: Supplementary Material

Julian Lienen, Eyke Hüllermeier

## A  Evaluation Details

### A.1  Dataset and Data Preprocessing

As datasets, we consider a wide range of commonly studied image classification datasets. While CIFAR-10 and -100 (Krizhevsky and Hinton, 2009), MNIST (LeCun et al., 1998), FashionMNIST (Xiao, Rasul, and Vollgraf, 2017) and SVHN (Netzer et al., 2011) (the latter three are used in an additional study in Sec. C.3) are widely known and well-studied, WebVision (Li et al., 2017) and Clothing1M (Xiao et al., 2015) comprise real-world noise from human annotations. For WebVision (version 1.0), we consider the Google image subset of ca. 66,000 images from the top-50 classes resized to 256x256. Clothing1M consists of 1 million training images with noisy and 10,000 test images with clean labels. Here, we also resize the images to 256x256.

For training, we apply data augmentation to the training images as commonly being done in image classification. To do so, we randomly crop images of size 32 (CIFAR-10(0)(N), SVHN), 227 (WebVision) or 224 (Clothing1M), followed by horizontally flipping the image with probability of 0.5. In case of MNIST and FashionMNIST, we keep the images unchanged, but resize them to size 32. The reported dimensions are used to preserve comparability with previous results (e.g., as reported in (Liu et al., 2020)).

### A.2  Synethtic Noise Model

While CIFAR-10N and -100N (Wei et al., 2022), WebVision and Clothing1M provide real-world noise in the standard annotations, we modeled additional label corruptions for the rest of the datasets in a synthetic manner. Thereby, we distinguish symmetric and asymmetric noise: For the former, each class is treated equally, whereas the asymmetric noise applies individual corruptions to each class.

In case of the symmetric noise, we flip a parameterized fraction $\rho \in [0, 1]$ of instances by uniformly sampling the label from all classes, i.e., we sample a corrupted label via $y \sim \mathrm{Unif}(\mathcal{Y})$ from the uniform distribution $\mathrm{Unif}(\mathcal{Y})$ over the class space $\mathcal{Y}$ with probability $\rho$. Hence, also correct label can be chosen again. In our studies, we considered values $\rho \in \{0.25, 0.5, 0.75\}$.

In case of asymmetric noise, we use the same setup as suggested in (Patrini et al., 2017). For CIFAR-10, we flip "truck" to "automobile", "bird" to "airplane", "deer" to "horse", as well as interchange "cat" and "dog". For CIFAR-100, the classes are grouped into 20 clusters with 5 classes each, whereby labels are flipped within these clusters in a round-robin fashion. In both cases, we apply asymmetric random flips with probability $\rho = 0.4$.

## A.3 Hyperparameters

|  | CIFAR-10(N) | CIFAR-100(N) | WebVision | Clothing1M |
|---|---|---|---|---|
| Model | ResNet34 | ResNet34 | ResNet50 (pretrained) | ResNet50 (pretrained) |
| Batch size | 128 | 128 | 64 | 32 |
| Learning rate (LR) | 0.02 | 0.02 | 0.02 | 0.002 |
| LR decay | cosine | cosine | cosine | cosine |
| Weight decay | $5 \times 10^{-4}$ | $5 \times 10^{-4}$ | $5 \times 10^{-4}$ | $1 \times 10^{-3}$ |
| Epochs | 120 | 150 | 100 | 15 |

Table 1: Hyperparameters fixed for all baselines and our method per dataset. ResNet50 models proceed from model weights pretrained on ImageNet.

In our paper, we follow common evaluation settings as in recent label noise robustness approaches (Zhou et al., 2021; Liu et al., 2022). To this end, we keep the optimizer hyperparameters as being used in previous reports, allowing for a fair comparison. Table 1 gives an overview over used hyperparameters. In all case, we used SGD with momentum of 0.9 as optimizer.

For the individual losses, we used the optimal parameters being reported in the respective works. This is reasonable as these parameters were optimized in a similar manner on the same data, such that we can re-use these optimization results. For our losses, we fixed the parameters as specified above and optimized the $\beta$ parameters using random search with 20 iterations employing a 5-fold cross validation on the (partially noisy) training data. To ensure this optimization does not give our method an unfair advantage, we performed the same operation to samples of the baseline experiments, where we noticed that the optimization preferred similar hyperparameter combinations as also reported in the related work.

For results presented in the main paper, we used a cosine decaying strategy as shown in Eq. (5) with $\beta_0 = 0.75$ and $\beta_1 = 0.6$ on all datasets except CIFAR-100. On the latter, we used a fixed schedule with $\beta = 0.5$. In Section C.2, we further present an ablation study showing results for different $\beta$ schedules. In all cases, we set $\alpha = 0.05$.

## A.4 Technical Infrastructure

The code of our empirical evaluation is publicly availabe[1]. All experiments are implemented using PyTorch[2] as deep learning framework using Nvidia CUDA[3] as acceleration backend. Image manipulations as being used for data agumentation are provided by Pillow[4]. To perform the experiments, we used several Nvidia A100 and RTX 2080 Ti accelerators in a modern high performance computing environment.

## B Experimental Result Completion

Due to space limitations, some results of the experiments in the main paper have been shifted to this section. The following subsections complete the presentation of the results.

### B.1 Synthetic Noise Results

| Loss | Add. Param. | CIFAR-10 | | | | CIFAR-100 | | | |
|---|---|---|---|---|---|---|---|---|---|
| | | | Sym. | | Asym. | | Sym. | | Asym. |
| | | 25 % | 50 % | 75 % | 40 % | 25 % | 50 % | 75 % | 40 % |
| CE | ✗ | 79.05 ±0.67 | 55.03 ±1.02 | 30.03 ±0.74 | 77.90 ±0.31 | 58.27 ±0.36 | 37.16 ±0.46 | 13.66 ±0.45 | 62.83 ±0.25 |
| LS ($\alpha = 0.1$) | ✗ | 76.66 ±0.69 | 53.95 ±1.47 | 29.03 ±1.21 | 78.07 ±0.69 | 59.75 ±0.24 | 37.61 ±0.61 | 13.53 ±0.51 | 63.76 ±0.51 |
| LS ($\alpha = 0.25$) | ✗ | 77.48 ±0.32 | 53.08 ±1.95 | 28.29 ±0.65 | 77.35 ±0.76 | 59.84 ±0.57 | 39.80 ±0.38 | 14.18 ±0.44 | 63.33 ±0.25 |
| LR ($\alpha = 0.1$) | ✗ | 80.53 ±0.39 | 57.55 ±0.95 | 29.83 ±0.87 | 77.83 ±0.37 | 57.52 ±0.58 | 36.77 ±0.54 | 13.23 ±0.14 | 62.46 ±0.15 |
| LR ($\alpha = 0.25$) | ✗ | 80.43 ±0.09 | 60.18 ±1.01 | 31.36 ±0.91 | 78.35 ±0.72 | 57.67 ±0.11 | 37.15 ±0.14 | 13.41 ±0.24 | 62.85 ±0.53 |
| GCE | ✗ | 90.82 ±0.10 | 83.36 ±0.65 | 54.34 ±0.37 | 77.37 ±0.94 | 68.06 ±0.31 | 58.66 ±0.28 | **26.85** ±1.28 | 61.08 ±0.51 |
| NCE | ✗ | 79.05 ±0.12 | 63.94 ±1.74 | 38.23 ±2.63 | 76.84 ±0.41 | 19.32 ±0.81 | 11.09 ±1.03 | 6.12 ±7.57 | 24.67 ±0.71 |
| NCE+AGCE | ✗ | 87.57 ±0.10 | 83.05 ±0.81 | 51.16 ±6.44 | 69.75 ±2.33 | 64.15 ±0.23 | 39.64 ±1.66 | 7.67 ±1.25 | 53.87 ±1.60 |
| NCE+AUL | ✗ | 88.89 ±0.29 | 84.18 ±0.42 | **65.98** ±1.56 | 80.87 ±0.34 | 69.76 ±0.31 | 57.41 ±0.41 | 17.72 ±1.27 | 61.33 ±0.55 |
| CORES | ✗ | 88.60 ±0.28 | 82.44 ±0.29 | 47.32 ±17.03 | 82.22 ±0.55 | 60.36 ±0.67 | 46.01 ±0.44 | 18.23 ±0.28 | 65.06 ±0.41 |
| RDA (ours) | ✗ | **91.48** ±0.22 | **86.47** ±0.42 | 48.11 ±15.41 | **85.95** ±0.40 | **70.03** ±0.32 | **59.83** ±1.15 | 26.75 ±8.83 | **69.62** ±0.54 |
| ELR | ✓ | 92.45 ±0.08 | 88.39 ±0.36 | 72.58 ±1.63 | 82.18 ±0.42 | <u>73.66</u> ±1.87 | 48.72 ±26.93 | 38.35 ±10.26 | <u>74.19</u> ±0.23 |
| SOP | ✓ | <u>92.58</u> ±0.08 | <u>89.21</u> ±0.33 | <u>76.16</u> ±4.88 | 84.61 ±0.97 | 72.04 ±0.67 | <u>64.28</u> ±1.44 | <u>40.59</u> ±1.62 | 64.27 ±0.34 |

Table 2: Test accuracies and standard deviations on the test split for models trained on CIFAR-10(0) with synthetic noise (symmetric or asymmetric). The results are averaged over runs with different seeds, bold entries mark the best method without any additional model parameters. Underlined results indicate the best method overall.

---

[1] https://github.com/julilien/MitigatingLabelNoiseDataAmbiguation
[2] https://pytorch.org/, BSD-3 license
[3] https://developer.nvidia.com/cuda-toolkit
[4] https://python-pillow.org/, HPND License

## B.2 Complete Real-World Noise Results

| Loss | Add. Param. | CIFAR-10N | | | | | | CIFAR-100N | |
|---|---|---|---|---|---|---|---|---|---|
| | | Clean | Random 1 | Random 2 | Random 3 | Aggregate | Worst | Clean | Noisy |
| CE | ✗ | **94.12** ±0.17 | 82.96 ±0.23 | 83.16 ±0.52 | 83.49 ±0.34 | 88.74 ±0.13 | 64.93 ±0.79 | 75.29 ±0.15 | 52.88 ±0.14 |
| LS ($\alpha = 0.1$) | ✗ | 93.92 ±0.03 | 82.76 ±0.47 | 82.10 ±0.21 | 82.12 ±0.37 | 88.63 ±0.11 | 63.10 ±0.38 | 75.71 ±0.19 | 53.48 ±0.45 |
| LS ($\alpha = 0.25$) | ✗ | 93.71 ±0.40 | 82.95 ±1.57 | 83.86 ±2.05 | 82.61 ±0.25 | 87.03 ±2.29 | 66.14 ±6.89 | 75.69 ±0.17 | 53.98 ±0.27 |
| LR ($\alpha = 0.1$) | ✗ | 93.77 ±0.06 | 83.00 ±0.36 | 82.64 ±0.31 | 82.82 ±0.21 | 88.41 ±0.29 | 66.62 ±0.33 | 74.79 ±0.16 | 52.01 ±0.04 |
| LR ($\alpha = 0.25$) | ✗ | 93.63 ±0.06 | 82.14 ±0.49 | 81.87 ±0.34 | 82.46 ±0.11 | 88.07 ±0.45 | 66.44 ±0.14 | 74.51 ±0.17 | 52.22 ±0.29 |
| GCE | ✗ | 93.22 ±0.08 | 88.85 ±0.19 | 88.96 ±0.32 | 88.73 ±0.11 | 90.85 ±0.32 | 77.24 ±0.47 | 72.29 ±0.19 | 55.43 ±0.47 |
| NCE | ✗ | 87.67 ±0.25 | 81.88 ±0.27 | 81.02 ±0.32 | 81.48 ±0.13 | 84.62 ±0.49 | 69.40 ±0.10 | 32.31 ±0.31 | 21.12 ±0.67 |
| NCE+AGCE | ✗ | 92.56 ±0.07 | 89.48 ±0.28 | 88.95 ±0.10 | 89.25 ±0.29 | 90.65 ±0.44 | 81.27 ±0.44 | 72.00 ±0.09 | 51.42 ±0.65 |
| NCE+AUL | ✗ | 93.09 ±0.10 | 89.42 ±0.22 | 89.36 ±0.15 | 88.94 ±0.55 | 90.92 ±0.19 | 81.28 ±0.47 | 74.18 ±0.21 | 56.58 ±0.41 |
| CORES | ✗ | 93.09 ±0.08 | 86.09 ±0.57 | 86.48 ±0.27 | 86.02 ±0.22 | 89.23 ±0.10 | 76.80 ±0.96 | 73.70 ±0.17 | 53.04 ±0.29 |
| RDA (ours) | ✗ | 94.09 ±0.19 | **90.43** ±0.03 | **90.09** ±0.29 | **90.40** ±0.01 | **91.71** ±0.38 | **82.91** ±0.83 | **76.21** ±0.64 | **59.22** ±0.26 |
| ELR | ✓ | <u>94.21</u> ±0.11 | <u>91.35</u> ±0.29 | <u>91.46</u> ±0.29 | <u>91.39</u> ±0.03 | <u>92.68</u> ±0.03 | <u>84.82</u> ±0.42 | <u>76.66</u> ±0.11 | <u>62.80</u> ±0.27 |
| SOP | ✓ | 92.84 ±0.20 | 89.16 ±0.40 | 89.02 ±0.33 | 88.99 ±0.31 | 90.54 ±0.16 | 80.65 ±0.13 | 76.12 ±0.32 | 59.32 ±0.41 |

Table 3: Test accuracies and standard deviations on the test split for models trained on CIFAR-10(0)N without any noise (clean) and real-world noise. The results are averaged over runs with different seeds, bold entries mark the best method without any additional model parameters. Underlined results indicate the best method overall.

## C Further Experiments

### C.1 Comparison to Full Ambiguation

In this section, we compare our method RDA to a full ambiguation adaptation: For prediction $\widehat{p}$ exceeding $\beta$ for a class different to the training label, we ambiguate the target information by a completely agnostic credal sets. Hence, this instance is completely ignored in the loss optimization. Table 4 shows the respective results, where significant improvements over the complete ambiguation can be observed for our method. Hence, incorporating credal sets with two fully plausible classes appears reasonable from a loss minimization context.

| CIFAR-10N | RDA | Complete Ambig. |
|---|---|---|
| Clean | **94.09** ±0.19 | 93.77 ±0.37 |
| Random1 | **90.43** ±0.03 | 87.04 ±0.18 |
| Random2 | **90.09** ±0.29 | 89.36 ±0.16 |
| Random3 | **90.40** ±0.01 | 89.04 ±0.21 |
| Aggregate | **91.71** ±0.38 | 88.73 ±0.64 |
| Worst | **82.91** ±0.83 | 76.57 ±0.91 |

Table 4: Averaged test accuracies and standard deviations computed over three runs with different seeds. **Bold** entries mark the best method.

## C.2 Parameter Ablation

### C.2.1 Schedules

In our paper, we proposed to use a cosine decaying strategy to determine $\beta$, reflecting an increase in the certainty of the model over the course of the training. Here, we compare this strategy to two alternative schedules, namely a constant and a linear function. Table 5 shows the respective results, where the more sophisticated schedules appear to be superior compared to the constant function.

| Schedule | CIFAR-10 | | |
| :---: | :---: | :---: | :---: |
| | 25 % | 50 % | 75 % |
| Constant ($\beta = 0.75$) | 90.72 ±0.44 | 83.97 ±0.81 | 54.33 ±11.70 |
| Linear ($\beta_0 = 0.75$, $\beta_1 = 0.6$) | 91.35 ±0.29 | 85.16 ±1.48 | 35.50 ±1.52 |
| Cosine ($\beta_0 = 0.75$, $\beta_1 = 0.6$) | **91.41** ±0.27 | **86.46** ±0.47 | **56.83** ±1.71 |

Table 5: Averaged test accuracies and standard deviations computed over three runs with different seeds. **Bold** entries mark the best method.

### C.2.2 Varying $\beta_0$ and $\beta_1$

| $\beta_1$ | CIFAR-10 | |
| :---: | :---: | :---: |
| | 25 % | 50 % |
| 0.75 | 90.93 | 83.74 |
| 0.6 | 91.32 | 84.21 |
| 0.5 | **91.65** | **85.98** |
| 0.4 | 85.72 | 66.11 |
| 0.3 | 81.89 | 41.93 |
| 0.2 | 78.49 | 11.79 |

Table 6: Results for different $\beta_1$ parameters with $\beta_0 = 0.75$ using a cosine decaying $\beta$ schedule. **Bold** entries mark the best parameter.

In another study, we look how changes in $\beta_0$ and $\beta_1$ for the cosine schedule behave in terms of generalization performance. The respective results are shown in Table 6 and 7.

## C.3 Simplified Setting

Relating to the phenomenon of neural collapse (Papyan, Han, and Donoho, 2020), we study the effects of our method in a simplified setting. More precisely, we consider multi-layer perceptron models consisting of a feature encoder and a classification head. The models consist of 6 dense layers with a width of 2048 neurons. To conform with previous studies (Nguyen et al., 2023), we restrict

| $\beta_0$ | CIFAR-10 | |
|---|---|---|
|  | 25 % | 50 % |
| 0.8 | **90.18** | **82.63** |
| 0.75 | 89.41 | 81.71 |
| 0.7 | 88.17 | 80.58 |
| 0.6 | 86.14 | 78.44 |
| 0.5 | 69.20 | 65.71 |

Table 7: Results for different $\beta_0$ parameters with $\beta_1 = 0.5$ using a cosine decaying $\beta$ schedule. **Bold** entries mark the best parameter.

the number of features at the last encoder layer to the number of classes $K$ in the datasets. We consider the datasets MNIST, FashionMNIST and SVHN, from which we sample classes $K \in \{2, 3, 5, 10\}$. Apart from the model, we use the same hyperparameters as described in Table 1 for CIFAR-10. We used the same setup to construct Figure 3 as shown in the main paper. For $K = 2$, we can readily investigate the learned feature representations over the course of the training in a convenient manner.

These experiments aim to provide further evidence for the generalization of our method in more restricted settings under simplified model assumptions. Table 8 shows the results, where a consistent improvement of our method compared to typically considered neural collapsing functions can be observed. This suggests that our method is also applicable for more restricted models than typically considered overparameterized models.

| Loss | MNIST | | | | FashionMNIST | | | | SVHN | | | |
|---|---|---|---|---|---|---|---|---|---|---|---|---|
|  | $K=2$ | $K=3$ | $K=5$ | $K=10$ | $K=2$ | $K=3$ | $K=5$ | $K=10$ | $K=2$ | $K=3$ | $K=5$ | $K=10$ |
| CE | 73.19 | 67.59 | 68.44 | 66.24 | 78.60 | 68.33 | 65.68 | 59.51 | 76.60 | 65.65 | 58.59 | 50.57 |
| LS ($\alpha = 0.1$) | 74.04 | 66.44 | 72.19 | 71.24 | 81.45 | 72.23 | 67.60 | 62.78 | 79.66 | 68.97 | 61.74 | 52.72 |
| LR ($\alpha = 0.1$) | 73.66 | 66.00 | 66.34 | 64.58 | 80.15 | 69.43 | 64.20 | 57.25 | 76.63 | 67.00 | 58.62 | 49.98 |
| RDA (ours) | **97.59** | **90.78** | **87.86** | **82.75** | **92.45** | **91.63** | **82.36** | **77.13** | **86.77** | **73.03** | **65.67** | **53.98** |

Table 8: Test accuracies on the three additional benchmark datasets using a simplified MLP with a restricted feature space at the penultimate layer. **Bold** entries mark the best method.

## C.4 Combination with Sample Selection

In additional experiments, we integrated our RDA approach in a sample selection approach based on the small loss criterion (Gui, Wang, and Tian, 2021), as also been applied in more sophisticated (semi-supervised) approaches that add substantial complexity to the learning setup. To this end, we train the model for 3 epochs on CIFAR-10(N)/-100 with cross-entropy based on the training examples, and take the 10 % training instances with the smallest cross-entropy loss. For these instances, we fix the label, i.e., we do not allow any ambiguity

such that these instances serve as a corrective. Then, we train the model with our RDA loss as described in Algorithm 1 with the hyperparameters reported in Tab. 1. In the following, we will refer to this variant as *RDA\**.

| Method | CIFAR-10 | | | CIFAR-100 | | |
|---|---|---|---|---|---|---|
| | 25 % | 50 % | 75 % | 25 % | 50 % | 75 % |
| RDA | 91.48 ±0.22 | 86.47 ±0.42 | 48.11 ±15.41 | **70.03** ±0.32 | 59.83 ±1.15 | 26.75 ±8.83 |
| RDA* | **91.79** ±0.21 | **86.78** ±0.50 | **67.08** ±2.09 | 69.98 ±0.17 | **60.18** ±1.18 | **30.82** ±9.45 |

Table 9: Test accuracies and standard deviations on the test split for models trained on CIFAR-10(0) with synthetic noise. The results are averaged over runs with different seeds, **bold** entries mark the best performing method per column.

| Method | CIFAR-10N | | | | | |
|---|---|---|---|---|---|---|
| | Clean | Random 1 | Random 2 | Random 3 | Aggregate | Worst |
| RDA | 94.09 ±0.19 | 90.43 ±0.03 | 90.09 ±0.29 | 90.40 ±0.01 | 91.71 ±0.38 | 82.91 ±0.83 |
| RDA* | **94.15** ±0.05 | **90.76** ±0.13 | **90.55** ±0.37 | **90.86** ±0.09 | **91.84** ±0.18 | **83.79** ±0.24 |

Table 10: Test accuracies and standard deviations on the test split for models trained on CIFAR-10N with synthetic noise. The results are averaged over runs with different seeds, **bold** entries mark the best performing method per column.

As can be seen in the results present in Tables 9 and 10, RDA* can achieve almost consistently better performance with this addition, confirming its flexibility in being employed in more sophisticated setups. Notably, RDA* shows the largest improvement in robustness in the high noise regime (75 % noise).

# D  Additional Training Behavior Plots

## D.1  CIFAR-10

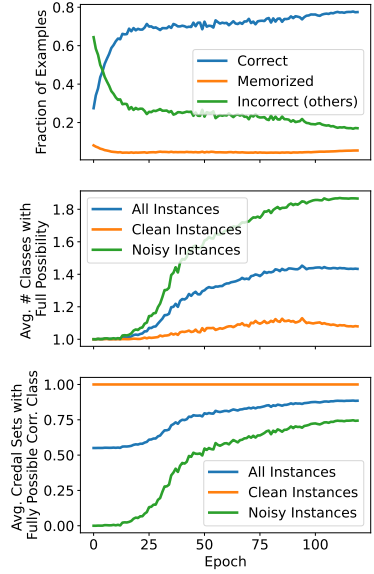

Figure 1: The training behavior of our method on CIFAR-10 with 50 % label noise averaged over five different seeds.

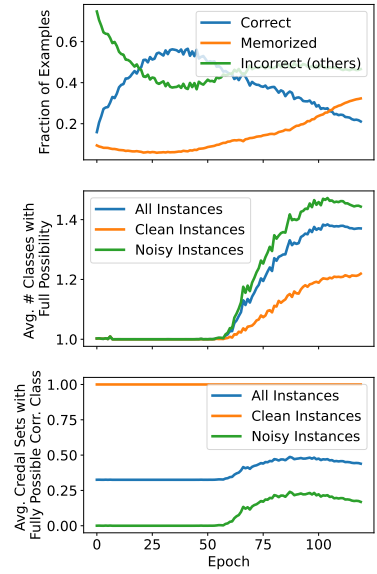

Figure 2: The training behavior of our method on CIFAR-10 with 75 % label noise averaged over five different seeds.

## D.2 CIFAR-100

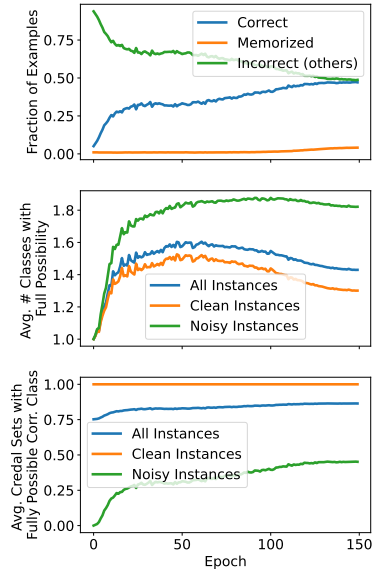

Figure 3: The training behavior of our method on CIFAR-100 with 25 % label noise averaged over five different seeds.

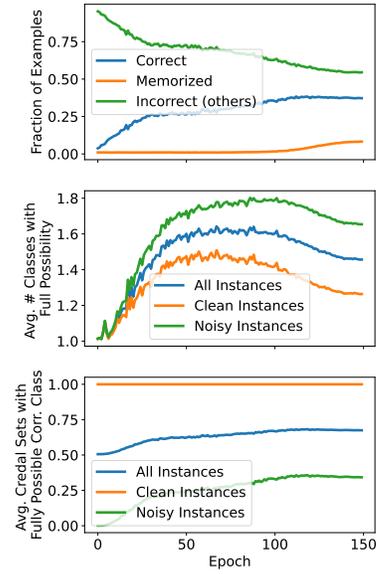

Figure 4: The training behavior of our method on CIFAR-100 with 50 % label noise averaged over five different seeds.

9

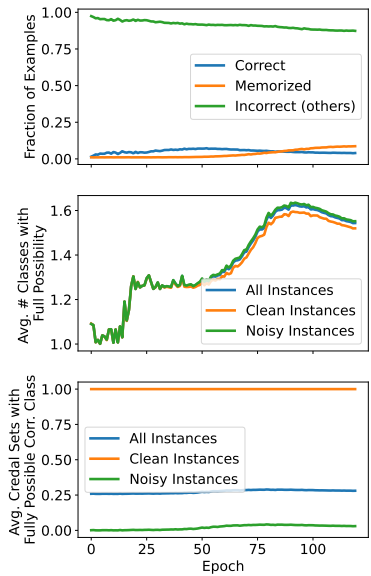

Figure 5: The training behavior of our method on CIFAR-100 with 75 % label noise averaged over five different seeds.



\end{document}